\documentclass{article}
\usepackage[T1]{fontenc}
\usepackage{microtype,graphicx,booktabs}
\usepackage{amsmath,amssymb,amsthm}
\usepackage{hyperref}
\ifdefined\anonymousversion
  \usepackage{icml2026}
\else
  \usepackage[preprint]{icml2026}
\fi
\hypersetup{hidelinks,pdftitle={NumericJev: Jev-like LLM Numerical Decoding with Multiway Decision Trees}}
\ifdefined\anonymousversion
  \hypersetup{pdfauthor={Anonymous Authors}}
\else
  \hypersetup{pdfauthor={WEIWEI YE, HANGCHEN LIU, and RENHE JIANG}}
\fi
\newcommand{\method}{\mbox{\textsc{NumericJev}}}
\newcommand{\E}{\mathbb{E}}

\newtheorem{proposition}{Proposition}
\icmltitlerunning{Numerical Decoding with Multiway Decision Trees}
\makeatletter
\ifdefined\anonymousversion
  \gdef\icmlcorrespondingauthor@text{Anonymous Author}
\else
  \gdef\icmlcorrespondingauthor@text{RENHE JIANG}
\fi
\makeatother

\begin{document}
\ifdefined\anonymousversion
  \hypersetup{pdfsubject={Anonymous research manuscript}}
\else
  \hypersetup{pdfsubject={Research preprint}}
\fi
\twocolumn[\icmltitle{NumericJev: Jev-like LLM Numerical Decoding\\
with Multiway Decision Trees}
\begin{icmlauthorlist}
\icmlauthor{WEIWEI YE}{utokyo}
\icmlauthor{HANGCHEN LIU}{utokyo}
\icmlauthor{RENHE JIANG}{utokyo}
\end{icmlauthorlist}
\icmlaffiliation{utokyo}{The University of Tokyo}
\icmlkeywords{numerical decoding, structured decisions, language models, hierarchical search}
\vskip 0.3in]
\printAffiliationsAndNotice{}
\hypersetup{pdftitle={NumericJev: Jev-like LLM Numerical Decoding with Multiway Decision Trees}}

\begin{abstract}
Large language models can interpret natural language, yet robust decisions remain
challenging. Jev-like models expose structured choices, but these interfaces do not
directly provide numerical values at a requested precision. We propose \method{},
a training-free numerical decoding algorithm that enables numerical output from any
LLM with a Jev-like structured-choice interface. Surprisingly, on our arithmetic
benchmark, it outperforms direct selection from a candidate list containing the
correct answer by 2.93 percentage points (Figure~\ref{fig:performance}). Our motivation comes from the observation that numerical range selection
is itself a decision problem that Jev-like LLMs can address. \method{} recursively
refines a range through a multiway decision tree while retaining the original
question in context, without parameter updates or hidden-state access. On a 100-value
grid, a ten-way tree requires only two decision rounds. Range-normalized MAE is
1.84\% versus 5.18\% for direct choice. A separate three-date historical-index
study yields 4.58\% mean relative recall error and 0\% readout error when the value
is supplied.
\ifdefined\anonymousversion
\else
Code is available at \url{https://github.com/Bring-AI/jev-numeric}.
\fi
\end{abstract}

\begin{figure*}[t]
\centering
\includegraphics[width=\textwidth]{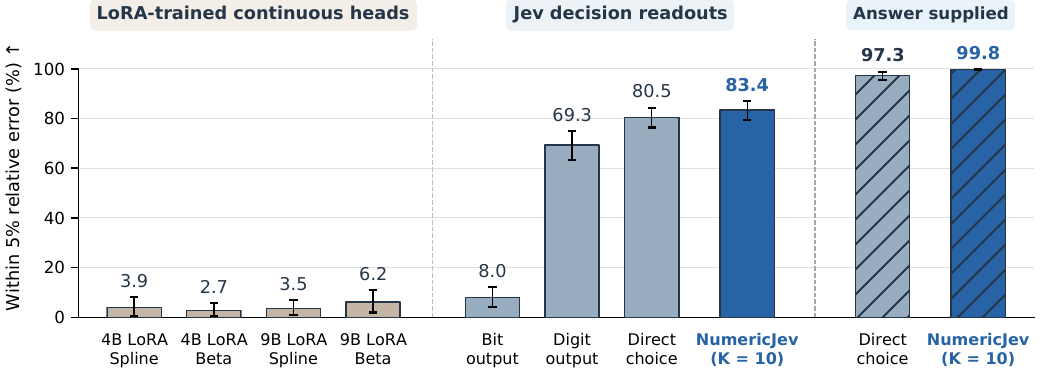}
\caption{\textbf{Numerical output through different interfaces.} Percentage of predictions
within 5\% relative error on all 256 arithmetic expressions; error bars give 95\% family-bootstrap
intervals. Continuous heads use 4B/9B backbones with LoRA \citep{hu2021lora}, trained on structural-causal
(SCM) distributions and transferred without arithmetic training; their means are mapped
back from $[0,1]$. Jev readouts average two option orders. The hatched bars compare
direct choice and ten-way \method{} in the separate supplied-answer condition. This post-hoc
threshold metric supplements the primary MAPE results; backbone and training differences
prevent interpreting the comparison as an isolated output-head ablation.}
\label{fig:performance}
\end{figure*}

\section{Introduction}

Large language models (LLMs) can interpret natural-language instructions
\citep{ouyang2022instructions}, yet turning
this understanding into robust decisions remains difficult. Even changing the order
of otherwise identical answer options can alter their selections \citep{zheng2024choice}.
Decision-oriented models such as Jev expose explicit choices and option probabilities
\citep{jev2026choice}. We refer to this
interface class as \emph{Jev-like}. Its focus on decisions, however, leaves a practical
gap: categorical outputs and rubric scores do not directly provide arbitrary numerical
values at a requested precision \citep{jev2026choice,jev2026score}.

Jev-like models can make effective decisions among alternatives described in natural
language \citep{jev2026choice}, yet this strength does not directly translate into
reliable numerical values (Figure~\ref{fig:performance}).
Existing numerical-readout studies include training regression heads on hidden
representations \citep{piskorz2026eliciting} and learning mappings from numeral
embeddings to values \citep{wallace2019numeracy}. These formulations expose two distinct
concerns: numerical training adds a mapping that must generalize beyond its supervision,
while restricting inputs to isolated numbers removes the linguistic context needed to
interpret a full question \citep{wallace2019numeracy}. The same study finds that number-decoding probes
extrapolate poorly outside the training range, and that numerical reasoning degrades
when digits in passages are rewritten as words. In our separate transfer evaluation,
SCM-trained spline and Beta heads struggle on arithmetic questions, reaching only
2.73--6.25\% within 5\% relative error (Figure~\ref{fig:performance};
Appendix~\ref{app:transfer}). The challenge is
therefore to obtain numerical precision while retaining the existing model's ability
to interpret the question and its context.

\begin{figure*}[t]
\centering
\includegraphics[width=\textwidth]{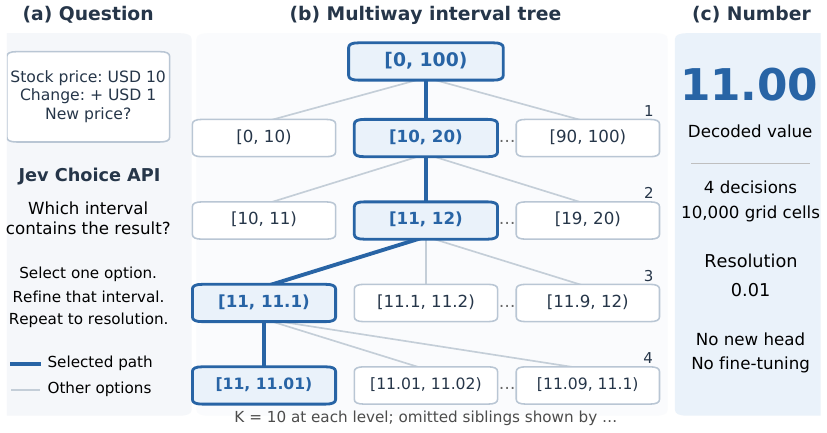}
\caption{\textbf{NumericJev as a multiway decision tree.} A recorded $10+1$ query
follows four ten-way decisions from $[0,100)$ to $[11,11.01)$. Each level partitions
only the selected parent; three siblings are displayed and the remaining alternatives
are elided. Thick edges show the selected path. The returned lower endpoint is 11.00;
the final cell specifies resolution, not confidence. No model parameters are updated.}
\label{fig:pipeline}
\end{figure*}

We investigate \method{}, a training-free numerical decoder for a Jev-like
structured-choice interface. Our starting point is that a numerical value can be
located to a chosen precision through successive decisions about where it lies,
consistent with reductions from numerical prediction to classification
\citep{torgo1997regression,langford2006quantiles}.
Selecting among candidate numerical ranges is itself a decision problem; we therefore
ask whether Jev-like LLMs can produce reliable numerical values through a sequence
of such choices.
Partitioning a range into intervals turns numerical estimation into a choice among
descriptions that the model can evaluate in the context of the original question.
Selecting an interval and repeating this construction progressively increases
resolution, provided the decisions retain the target (Proposition~\ref{prop:depth}).
The model supplies contextual
judgments; the external decoder handles interval arithmetic and numerical output.
This division of labor reuses the existing decision capability without fitting a new
mapping from hidden states to numbers (Figure~\ref{fig:pipeline}; Algorithm~\ref{alg:decode}).

Specifically, \method{} decodes a numerical value by organizing candidate values into
a multiway interval tree. At each level, Jev selects among $K$ subintervals in the
context of the original question; the decoder refines the selected interval and
returns the final cell's lower endpoint (Algorithm~\ref{alg:decode}). On our 100-value
evaluation grid, a ten-way tree ($K=10$) produces each numerical output in only two
iterations (Proposition~\ref{prop:depth}). Across all 256
arithmetic expressions and two option orders, 83.40\% of outputs fall within 5\%
relative error, compared with 80.47\% for direct choice
(Figure~\ref{fig:performance}).

Our contributions are: (1) We identify numerical range selection as a decision problem that naturally fits
the language-conditioned choice interface of Jev-like LLMs, and investigate whether
composing such decisions can yield reliable numerical outputs (Section~\ref{sec:method}).
(2) We propose \method{}, a training-free numerical decoder that recursively
refines a range through a multiway decision tree, without parameter updates or
hidden-state access. On a 100-value grid, a ten-way tree ($K=10$) requires only two
decision rounds to produce a numerical output (Algorithm~\ref{alg:decode};
Proposition~\ref{prop:depth}).
(3) On 256 arithmetic expressions, ten-way \method{} places 83.40\% of outputs
within 5\% relative error and reduces range-normalized MAE from 5.18\% for direct
choice to 1.84\% (Figure~\ref{fig:performance}; Table~\ref{tab:main}). A separate
three-date historical-index study yields 4.58\% mean relative recall error and 0\%
supplied-value readout error (Table~\ref{tab:history}).

\ifdefined\anonymousversion
Implementation, evaluation records, and generation scripts accompany the supplementary
artifact. The named repository link is omitted for anonymous review.
\fi

\section{Related Work}

\paragraph{Regression through classification.}
Discretizing a continuous target and applying classification predates modern language
models. \citet{torgo1997regression} study interval discretization and misclassification
costs for regression, and \citet{langford2006quantiles} reduce conditional quantile
prediction to classification with a regret analysis. These approaches establish that
classification can support numerical prediction. Our setting differs in its access
constraints: the classifier is a fixed hosted choice interface, the alternatives can
change at inference time, and no classifier or numerical head is trained. Our analysis
describes the outer decoder rather than providing a learning-theoretic guarantee for
the underlying model.

\paragraph{Numerical outputs from language models.}
Numerical language-model interfaces include digit serialization for time series
\citep{gruver2023forecast} and language-conditioned predictive distributions
\citep{requeima2024llm}. Numerical tasks also require a decision rule appropriate for
the evaluation loss: a likely textual answer need not minimize numerical error
\citep{lukasik2024regression}. These results motivate evaluating numerical distance
and keeping a point decision distinct from a probability distribution. Our decoder
does not request a free-form numerical completion; its external interface consists of
successive structured choices. We make no claim about the provider's internal decoding
implementation.

\paragraph{Representation probes and choice sensitivity.}
Learned probes can estimate numerical distributional summaries from language-model
representations \citep{piskorz2026eliciting}. They require internal representations and
training examples, whereas our decoder uses the existing choice endpoint. Meanwhile,
multiple-choice behavior can depend on option identity and order
\citep{zheng2024choice}. This motivates explicit ordering controls but does not identify
the cause of any bias in Jev. The Jev documentation already describes chaining choice
questions for hierarchical classification \citep{jev2026choice}; our focus is the
resulting numerical error, including its dependence on depth and representation.

\section{Numerical Decoding through Structured Decisions}
\label{sec:method}

\subsection{Interface and finite numerical grid}

Let $x$ denote the input state and $y\in[L,U)$ the numerical target described by a
question. The choice interface accepts an instruction $q$ and a finite set of labeled
criteria $C$, returning a label $c$ and probabilities $p(c\mid x,q,C)$. We use the
returned selected label. The model's categorical probabilities are not assumed to be
calibrated numerical uncertainty. Native rubric scores are also distinct from an
arbitrary target value: our addition is a caller-specified numerical range and precision.

Choose a resolution $\epsilon>0$ such that $N=(U-L)/\epsilon$ is an integer. The grid
contains $N$ cells $[L+i\epsilon,L+(i+1)\epsilon)$, $i=0,\ldots,N-1$. The decoder
returns the lower endpoint of a selected cell. Its state is an interval of integer
indices $[a,b)$, initially $[0,N)$. Integer indices and decimal arithmetic avoid
floating-point ambiguity at boundaries. The user supplies both the containing range
and the stopping resolution; we do not infer a range or provide an escape branch.

\subsection{Balanced multiway refinement}

At each step, let $k=\min(K,b-a)$, where $K\geq2$ is the branching factor. Define
\begin{equation}
 t_j=a+\left\lfloor\frac{(b-a)j}{k}\right\rfloor,
 \quad j=0,\ldots,k.
\end{equation}
The criteria describe the numerical intervals
$[L+\epsilon t_j,L+\epsilon t_{j+1})$. A choice of label $j$ sets
$(a,b)\leftarrow(t_j,t_{j+1})$. Refinement stops when $b-a=1$, returning
$\hat y=L+\epsilon a$. Partitions differ by at most one cell, so the method also
handles grid sizes that are not powers of $K$.

\begin{algorithm}[t]
\caption{\method{}: finite-grid interval decoding}
\label{alg:decode}
\begin{algorithmic}[1]
\REQUIRE State $x$, target question, range $[L,U)$, resolution $\epsilon$, branching $K$
\STATE $a\gets0$, $b\gets (U-L)/\epsilon$
\WHILE{$b-a>1$}
\STATE $k\gets\min(K,b-a)$
\STATE $t_j\gets a+\lfloor(b-a)j/k\rfloor$, $j=0,\ldots,k$
\STATE Describe $C_j=[L+\epsilon t_j,L+\epsilon t_{j+1})$
\STATE $j^\star\gets\mathrm{Choice}(x,\text{target question},C)$
\STATE $(a,b)\gets(t_{j^\star},t_{j^\star+1})$
\ENDWHILE
\STATE \textbf{return} $\hat y=L+\epsilon a$, cell $[\hat y,\hat y+\epsilon)$, decision trace
\end{algorithmic}
\end{algorithm}

\subsection{Resolution, depth, and error propagation}

\begin{proposition}[Depth and exact-oracle readout]
\label{prop:depth}
For $N$ grid cells and branching $K\geq2$, the algorithm uses at most
$D=\lceil\log_K N\rceil$ sequential choices. If every selected interval contains $y$,
then $0\leq y-\hat y<\epsilon$. If $y$ is a grid value, $\hat y=y$.
\end{proposition}

Each chosen interval has at most $\lceil n/K\rceil$ cells when its parent has $n$
cells. Repeating this bound gives the stated depth. The numerical guarantee is
conditional on the selected path, not on the confidence returned by the model.
Smaller resolution alone therefore says little about practical accuracy.

\begin{proposition}[First-divergence error bound]
Let $E_d$ be the event that the first selected interval excluding $y$ occurs at depth
$d$. Define $W_{d-1}=\epsilon\lceil N/K^{d-1}\rceil$. Then
\begin{equation}
 \E|\hat y-y|\leq \epsilon+\sum_{d=1}^{D}W_{d-1}\Pr(E_d).
 \label{eq:error}
\end{equation}
No independence assumption between decisions is required.
\end{proposition}

Before the first divergence, the target and the eventual decoded value belong to the
same parent interval. Their distance is bounded by its width; the no-divergence event
contributes at most $\epsilon$. Early errors can consequently dominate later decimal
precision. For example, a target 0.49 lies close to an incorrectly selected interval
$[0.5,1)$, whose lower endpoint has error 0.01. Refining that interval to $[0.75,1)$
increases the lower-endpoint error to 0.26. Greedy refinement need not improve error
monotonically. Appendix~\ref{app:proofs} gives full proofs and the precise scope of the
bound.

\subsection{Alternative numerical representations}

Direct selection lists all grid values in one question. Sequential index-digit
decoding expresses the grid index $q=(y-L)/\epsilon$ as decimal digits, conditioning
each digit on the preceding selection. Independent index-bit decoding queries its
binary bits without such conditioning. They share the same stated in-range grid in
our controlled experiment, but ask different questions. A learned model must perform
the requested numerical comparison or representation conversion; mathematical
equivalence does not imply identical model behavior.

A selected tree path is only a point readout. To construct a distribution, a separate
experimental interface queries thresholds $p_i\approx P(Y\leq z_i\mid x)$, projects
them onto a nondecreasing sequence, and differences adjacent values into bin masses.
The projection can enforce a valid CDF shape but cannot establish calibration.
Section~\ref{sec:distribution} evaluates this distinction using archived threshold
queries. We do not interpret a product of probabilities along a greedily selected path
as a validated distribution over numerical outcomes.

\section{Experimental Design}
\label{sec:design}

\paragraph{Model and access.}
All choice-interface experiments use the hosted Jev version identified as \texttt{jev-1.13-20260917}.
Requests provide structured state and Choice questions. There are no parameter updates,
hidden-state accesses, external calculators, or retrieval tools supplied to the model.
Reference arithmetic is computed separately for evaluation. Multiple independent
questions are batched into one request; the service's internal inference, ordering,
and caching are not observable. We record the served model identifier and every
request and response.

\paragraph{Fixed affine-grid evaluation.}
We generate 64 distinct base expressions with seed 20260923: 16 each using addition,
subtraction, multiplication, and division. Every result $z$ is an integer in
$\{1,\ldots,99\}\setminus\{50\}$. Four transformations produce the targets
\begin{equation}
 y=z,\qquad y=0.01z,\qquad y=z-50,\qquad y=10000z.
\end{equation}
The corresponding $(L,\epsilon)$ pairs are $(0,1)$, $(0,0.01)$, $(-50,1)$,
and $(0,10000)$; $U=L+100\epsilon$ in every case. Thus there are 256 distinct
expressions but only 64 underlying families. Exact rational evaluation verifies every
target. Excluding zero makes the relative-error metric defined throughout; this
exclusion is a limitation for applications with zero-valued outcomes.

Each expression is evaluated in two conditions. \emph{Arithmetic} supplies only the
expression. \emph{Provided} supplies the identical expression and an additional field
containing its numerical result. Each condition uses ascending and reversed option
insertion order, preserving label-description pairs. The full design has 1,024
condition/order jobs and six outputs per job. Cases, prompts, model version, and job
order were fixed and hashed before execution; prompts were not revised after inspecting
these results. This controlled extension follows the exploratory experiments below,
so it is not a blind model-development benchmark or evidence of pretraining OOD.

\paragraph{Decoders.}
We compare direct selection among 100 grid values; interval refinement with $K=2,4,10$;
two sequential decimal index digits; and seven independently queried binary index
bits. Index encodings receive the same lower bound and step used by the other
methods. Their extra conversion burden is part of the tested interface, not a claim
about all possible digit prompts. Binary outputs above index 99 are retained without
clipping, so invalid codes cannot be hidden by postprocessing. Maximum dependent
rounds are seven, four, and two for the three trees, respectively; direct choice and
parallel bits require one round and decimal index digits require two. These are
dependency depths, not measurements of isolated method latency.

\paragraph{Numerical metrics and uncertainty.}
For nonzero references, absolute percentage error is
\begin{equation}
 \operatorname{APE}(\hat y,y)=100\frac{|\hat y-y|}{|y|}.
\end{equation}
Our primary endpoint, MAPE, averages APE equally over cases and orders.
We also report range-normalized mean absolute error,
$\operatorname{NMAE}=100\E[|\hat y-y|/(U-L)]$, which is invariant to the four
affine transforms for a fixed grid-index error. Near-zero targets can dominate MAPE;
NMAE answers the complementary question of how far predictions move across the
specified range. Neither metric is an exact-match count.

Confidence intervals use 10,000 percentile-bootstrap replicates with seed 20260923.
We resample the 64 base families and keep their four transformations, both orders,
and paired input conditions together. Comparisons against direct choice use the
same resampled families. Repeated conditions and transformed questions are not treated
as independent evidence. These intervals describe this expression generator and model
version, not a population of all numerical tasks.

\paragraph{Archived exploratory evaluations.}
The earlier arithmetic study contains 12 hand-picked questions, six integer and six
dyadic-fraction expressions, with two orders and two repeats. It compares seven
representations on a 16-value grid and additionally probes threshold distributions.
The historical study asks for the S\&P~500 price index's official closing level on the
last trading day of 2019, 2020, and 2023. Verified references are 3230.78, 3756.07, and
4769.83, respectively, from the recorded source audit \citep{fred2026sp500}.
Ten-way decoding starts from $[0,10000)$ at 0.01-point resolution. A matched control
adds the reference close to state. Each date has two orders and two repeats, giving
12 outputs per input condition but only three distinct dates. These questions test
historical recall rather than forecasting or demonstrated training-data membership.

\begin{table*}[t]
\caption{\textbf{Fixed affine-grid evaluation.} Relative errors in percent, lower is
better. Each condition contains 512 outputs per method: 64 families $\times$ four
transformations $\times$ two orders. Brackets give 95\% family-bootstrap intervals for
MAPE. NMAE divides absolute error by the supplied range width. Rounds denote sequential
dependencies, with independent questions batched.}
\label{tab:main}
\centering\small
\begin{tabular}{lrrrrr}
\toprule
& \multicolumn{2}{c}{MAPE ($\downarrow$)} & \multicolumn{2}{c}{NMAE ($\downarrow$)} & \\
\cmidrule(lr){2-3}\cmidrule(lr){4-5}
Decoder & Arithmetic & Provided & Arithmetic & Provided & Rounds \\
\midrule
Direct values & 21.68 [16.33, 27.69] & 3.79 [1.63, 6.30] & 5.18 & 1.37 & 1 \\
Tree ($K=2$) & 16.88 [9.13, 27.01] & 0.00 [0.00, 0.00] & 1.76 & 0.00 & 6--7 \\
Tree ($K=4$) & 17.25 [6.89, 31.48] & 1.27 [0.00, 3.81] & 1.60 & 0.03 & 3--4 \\
Tree ($K=10$) & 19.73 [9.62, 32.58] & 2.93 [0.00, 8.79] & 1.84 & 0.06 & 2 \\
Index digits & 84.43 [50.59, 127.63] & 59.82 [36.31, 87.97] & 9.92 & 5.42 & 2 \\
Index bits & 95.13 [73.04, 120.63] & 77.90 [62.97, 95.56] & 24.56 & 21.76 & 1 \\
\bottomrule
\end{tabular}
\end{table*}

\section{Results}

\subsection{Interval refinement and numerical distance}

Table~\ref{tab:main} compares representations on the same 100-cell grid. In the
arithmetic condition, interval trees have MAPE 16.88\%, 17.25\%, and 19.73\% for
$K=2,4,10$, respectively, compared with 21.68\% for direct selection. However,
the paired MAPE differences against direct selection have 95\% intervals
$[-14.48,7.03]$, $[-16.10,10.31]$, and $[-14.05,12.77]$ percentage points.
These data therefore do not establish lower population MAPE for any tree.
The same trees have NMAE 1.76\%, 1.60\%, and 1.84\%, compared with 5.18\% for
direct selection; their paired NMAE differences have intervals
$[-4.96,-2.02]$, $[-5.17,-2.13]$, and $[-4.87,-1.90]$ percentage points.
Both summaries are necessary: the relative ranking and uncertainty
depend on the numerical loss, particularly near zero.

Index digits and independent bits produce much larger gaps in this experiment:
84.43\% and 95.13\% MAPE, respectively. Their questions require converting the
computed result into a grid index and then into digits or bits. The finding concerns
these explicit encodings and prompts; it does not show that digit-based language
generation is inherently inferior. In particular, our index-digit baseline is not a
comparison against a freely generating language model or a learned numerical head.

\subsection{Supplying the value does not make every encoding equivalent}

The provided-value condition reduces MAPE to 3.79\% for direct selection,
0\% for $K=2$, 1.27\% for $K=4$, and 2.93\% for $K=10$.
For the binary tree, the paired difference from direct choice is $-3.79$ percentage
points with interval $[-6.30,-1.63]$. Zero observed error is confined to these tested
inputs and the selected decoder; it is not a guarantee of future performance.
Wider trees retain small range-normalized errors, 0.025\% and 0.059\%, which can
be amplified by a small reference magnitude in MAPE.

Providing the answer leaves index-digit and bit MAPE at 59.82\% and 77.90\%.
Thus the model can still fail at the requested representation conversion even when
the numerical target is explicit. Conversely, the arithmetic condition combines
computation and readout, so its improvement under answer provision cannot be attributed
exclusively to either source. The matched conditions expose this distinction without
requiring assumptions about hidden representations.

\subsection{Continuous-head transfer and a tolerance-based overview}

Figure~\ref{fig:performance} summarizes the full suite using the percentage of outputs
with $\operatorname{APE}\leq5\%$. This tolerance-based metric was added after the
original evaluation and does not replace its primary MAPE endpoint. Ten-way
\method{} reaches 83.40\%, compared with 8.01\% for bits, 69.34\% for index digits,
and 80.47\% for direct choice. With the answer supplied, direct choice reaches 97.27\%
and ten-way \method{} reaches 99.80\%.
Thus the large advantage over bit output coexists with a much smaller difference
from direct choice and a remaining 16.41-point gap to the answer-supplied control.

We additionally evaluate four existing LoRA-tuned continuous heads: spline and Beta
heads on 4B and 9B base backbones, trained on 12,000 structural-causal distribution
queries. They receive all 256 expressions, plus the known transform
$Y=(X-L)/(U-L)$ required by their $[0,1]$ support. We invert the predicted mean into
original units. No arithmetic training, calibration, or checkpoint selection on these
examples is performed. Their tolerance scores range from 2.73\% to 6.25\%.
The comparison measures transfer of these complete systems: Jev's backbone and training
are different, so it cannot establish that continuous heads are intrinsically inferior.
Appendix~\ref{app:transfer} records checkpoints, normalization, prompts, numerical
errors, and matched supplied-value controls.

\subsection{Scale, sign, and option ordering}

The affine design preserves the target grid index across transformations. An exact
choice oracle would therefore produce corresponding values, yet learned readout need
not be affine-equivariant. Table~\ref{tab:domains} reports the arithmetic condition
by domain. Its signed column contains targets close to zero, so it should be read
together with range-normalized errors rather than interpreted as an equally scaled
difficulty measure. The transformations also change the written expressions; this is
a controlled prompt-and-value transformation, not an isolated intervention on an
internal numerical representation.

Direct selection has 0\% MAPE on the unshifted integer domain, whereas the trees
have 0.50--3.19\%. On the large-scale domain, ten-way refinement has 30.96\% MAPE
versus 0.52\% for direct choice. These reversals rule out a uniformly preferred
branching factor and show why the aggregate cannot substitute for domain-level results.

We measure ordering sensitivity as the absolute difference between the two decoded
values divided by the range width. We also invert each affine transform and measure
the difference from the untransformed prediction on the shared 100-cell grid.
Appendix~\ref{app:sensitivity} reports both quantities. Reversing insertion order is
an external request control: it does not establish the order in which the backend
actually processes alternatives, nor test all permutations or relabelings.

\begin{table}[t]
\caption{\textbf{Arithmetic MAPE by affine domain (\%).} Each entry averages 64
families and two orders. The same underlying base results are transformed across
columns; these columns are not independent datasets.}
\label{tab:domains}
\centering\small
\begin{tabular}{lrrrr}
\toprule
Decoder & Integer & $\times0.01$ & $-50$ & $\times10^4$ \\
\midrule
Direct values & 0.00 & 2.72 & 83.47 & 0.52 \\
Tree ($K=2$) & 3.19 & 2.21 & 56.47 & 5.63 \\
Tree ($K=4$) & 0.78 & 2.39 & 63.38 & 2.42 \\
Tree ($K=10$) & 0.50 & 0.51 & 46.94 & 30.96 \\
Index digits & 1.21 & 13.07 & 298.93 & 24.53 \\
Index bits & 42.21 & 52.13 & 200.26 & 85.91 \\
\bottomrule
\end{tabular}
\end{table}

\subsection{Exploratory arithmetic and historical recall}

On the original 12-question arithmetic suite, four-way refinement has 2.42\% MAPE
and direct 16-value selection has 3.37\%. Other archived decoders have 8.48\% for
explicit-set membership, 11.29\% for parallel thresholds, 12.31\% for adaptive
thresholds, 15.85\% for decimal digits, and 25.27\% for independent bits.
These small-suite values cannot replace the broader fixed evaluation: the examples,
candidate grids, and prompts differ. Appendix~\ref{app:exploratory} preserves the
complete comparison and prompt definitions.

The historical recall experiment has 4.58\% mean, 4.83\% median, and 6.49\% maximum
relative error. The same protocol with supplied closing levels has 0\% for all three
summaries. Table~\ref{tab:history} reports distances per date. Every recall output is
above the reference. This directionality is an observation on three dates, not evidence
for a specific causal mechanism such as a token prior or a training-data bias.

\begin{table}[t]
\caption{\textbf{Historical index readout.} Absolute relative error across four
outputs per date; provided-value error is 0\% under this protocol. There are only
three unique dates.}
\label{tab:history}
\centering\small
\begin{tabular}{lrrr}
\toprule
Date & Reference & Recall APE & Provided APE \\
\midrule
2019-12-31 & 3230.78 & 3.11--4.96\% & 0\% \\
2020-12-31 & 3756.07 & 3.59--6.49\% & 0\% \\
2023-12-29 & 4769.83 & 4.83\% & 0\% \\
\midrule
Mean & & 4.58\% & 0\% \\
\bottomrule
\end{tabular}
\end{table}

A separately recorded follow-up changes both the prompt and the branching factor
to two. Recall MAPE rises to 31.17\% over six outputs, while supplied-value MAPE
remains 0\%. This is not a controlled branching-factor ablation and is not pooled
with the ten-way historical result. Together with the fixed-grid experiment, it
shows that the favorable historical readout result must remain attached to its exact
protocol.

\subsection{Numerical distributions need a separate validation}
\label{sec:distribution}

In the archived threshold experiment, 25 of 48 query groups have at least one
decrease in their proposed CDF. For one arithmetic input, the returned threshold
probabilities include $P(Y\leq0.4375)=0.26$ and $P(Y\leq0.5)=0.01$.
Equal-weight isotonic projection followed by adjacent differences yields nonnegative
histogram masses summing to one, given the prescribed endpoint masses. This is a
shape constraint. It neither recovers a unique ground-truth predictive distribution
nor validates calibration. Since the arithmetic references are deterministic, they
also do not supply a conditional outcome distribution for a forecasting study.
We therefore treat distribution construction as an unvalidated extension, rather than
as a consequence of the point-decoding results.

\section{Discussion and Limitations}

\method{} makes a numerical interface available through repeated choices, but its
resolution and its accuracy are separate properties. The number of representable
values can grow exponentially with the number of decisions, while an early incorrect
interval can remain unrecoverable. Wider trees reduce sequential depth without
guaranteeing lower error. The provided-value and affine controls show that the
representation requested at the interface is itself a meaningful source of variation.

The controlled choice-interface evaluation is limited to one hosted model version,
simple generated arithmetic, and a small historical recall example. The additional
continuous-head baselines use different backbones and prior training tasks; they do
not isolate the effect of the output head. We do not compare arithmetic-trained heads,
general-purpose text generation, open-ended word problems, or real forecasting datasets. Known bounds and resolutions are supplied by the evaluator; out-of-range
targets and unknown scales remain unsolved. Endpoint-valued arithmetic removes
quantization error and therefore evaluates decision errors more directly, but does
not establish performance on arbitrary real targets. The numerical digit and bit
prompts are only specific baselines. The bootstrap captures variation across the
sampled base expressions, not backend changes or all sources of model randomness.
The source contains all completed protocols, including unfavorable controls.

\section{Conclusion}

Numerical readout can be implemented over a structured-choice interface without
adding a trained numerical head. \method{} provides a concrete multiway interval
decoder, records every decision, and exposes its finite resolution. Controlled
experiments show smaller range-normalized errors than direct selection in the tested
arithmetic setting, but uncertain MAPE differences and sensitivity to scale, branching,
and representation. Supplying the value substantially changes the task without making
all encodings reliable. Numerical uncertainty and broader regression generalization
require additional evidence beyond a precise-looking decoded number.

\section*{Impact Statement}
This work studies numerical interfaces for language-model decisions. Its main risk is
that a fine decimal output or a narrow terminal interval may be mistaken for accuracy
or calibrated confidence. The historical index experiment is a recall diagnostic and
does not validate financial forecasting. The method should retain its decision trace,
state its imposed range and resolution, and be evaluated for the numerical losses and
failure costs of its intended application.

\bibliography{references}
\bibliographystyle{icml2026}

\clearpage
\appendix
\section{Proofs and Interpretation of the Error Bound}
\label{app:proofs}

\paragraph{Depth.}
For an interval with $n>K$ cells, each child has size at most $\lceil n/K\rceil$.
For $2\leq n\leq K$, every child is a singleton. Let $n_d$ denote the maximum size
after $d$ refinements along any path. Induction gives
$n_d\leq\lceil N/K^d\rceil$, using
$\lceil\lceil u\rceil/K\rceil=\lceil u/K\rceil$ for positive $u$ and integer $K$.
At $d=\lceil\log_K N\rceil$, this upper bound is one. If the path contains $y$ at
every step, the returned cell is exactly the unique half-open grid cell containing
$y$, which proves the endpoint error bound. The case $N=1$ needs no query.

\paragraph{First divergence.}
The events $E_1,\ldots,E_D$ are mutually exclusive. On $E_d$, all preceding
selected cells contained $y$. Both the final decoded value and $y$ lie in the parent
interval at depth $d-1$, even though the selected child excludes $y$. Their absolute
difference is less than that parent's width, at most
$W_{d-1}=\epsilon\lceil N/K^{d-1}\rceil$. On the complement of their union, the
absolute error is less than $\epsilon$. Conditioning on these disjoint events yields
the slightly sharper expression
\begin{equation}
\E|\hat y-y|
\leq \epsilon\Pr\!\left(\bigcap_{d=1}^{D}E_d^c\right)
  +\sum_{d=1}^{D}W_{d-1}\Pr(E_d),
\end{equation}
which implies Eq.~\ref{eq:error}. Different paths can terminate early; events beyond
termination are empty. This proof applies to in-range numerical targets and the
greedy decoder, not to unbounded outputs or targets outside the supplied range.

\paragraph{Conditional error probabilities.}
One can define $e_d$ as the probability of a divergence at depth $d$ conditional
on all earlier selected intervals containing the target. The chain rule, rather
than an independence assumption, gives
$\Pr(E_d)=e_d\prod_{j<d}(1-e_j)$ along the corresponding experiment.
This is not the product of the model's returned confidence scores. Establishing
that those scores estimate the required conditional probabilities would require
calibration evidence absent from this study.

\paragraph{Relative loss.}
If all reference magnitudes are bounded below by $m>0$, division by $m$ converts
the absolute-error bound into an upper bound on relative error. Without such a lower
bound, arbitrarily small references can magnify small absolute deviations. This is
why we report MAPE together with range-normalized error and do not interpret the
signed-domain MAPE as a scale-free estimate of difficulty.

\section{Dataset Generation and Exact Numerical Checks}

The generation script fixes seed 20260923 and samples operands until it obtains 16
distinct valid expressions per operator. For addition, both operands are drawn from
1 through 80. For subtraction, the first is drawn from 2 through 200 and the second
from 1 through 150. Multiplication operands are drawn from 2 through 12. Division
draws a denominator from 2 through 15 and an integer quotient from 1 through 99,
then forms the numerator as their product. All operators retain only integer results
strictly between 0 and 100, excluding 50. The operator order and seed are fixed in
the released script; rejection sampling is based solely on reference-domain validity,
not model outputs.

Each base expression $e$ is written as \texttt{(e) * s + (L)} for each of the four
domains. The result is verified by an independent restricted arithmetic AST evaluator
using exact rational numbers. The retained targets have no quantization error at their
specified resolution. The provided-value condition adds the result to the same state
without changing the target question. A deterministic exact-label oracle checks recovery
for all 256 cases, both option orders, and all six decoders before API execution.
This check verifies decoding mechanics, not model performance.

\section{Complete Prompt Specification}
\label{app:prompts}

All questions start with: ``Determine the numerical result x of the expression in
state.'' The base state contains only an \texttt{expression} field. The supplied-value
condition additionally contains a \texttt{result} field with a decimal string.

\paragraph{Direct choice.}
The instruction appends ``Select its numerical value.'' The criteria map string labels
0 through 99 to the 100 decimal grid values. Thus the option key denotes a grid index;
its description supplies the numerical candidate. No answer is embedded in the
arithmetic-only state or the question beyond the common complete candidate grid.

\paragraph{Interval trees.}
The appended instruction is: ``Select the interval containing x. Include the lower
bound and exclude the upper bound. Labels identify intervals, not numerical answers.''
Every criterion is written as \texttt{lower <= x < upper}, using the integer-grid
partition from Algorithm~\ref{alg:decode}. The prompt is identical for $K=2,4,10$;
only the number of intervals and the intervals determined by the selected path vary.

\paragraph{Decimal grid index.}
The prompt defines $q=(x-L)/\epsilon$ as an integer from 0 through 99. It asks for
exactly two decimal digits, including a leading zero if needed. The first question
selects the tens digit. The second supplies the previously selected tens digit and
selects the units digit. Both use criteria 0 through 9 with descriptions identifying
the requested digit. The decoded index is mapped back to $L+\epsilon q$.

\paragraph{Independent binary grid index.}
The prompt gives the same definition of $q$ and requests seven unsigned binary bits
with weights 64, 32, 16, 8, 4, 2, 1. For a bit of weight $w$, it explicitly states
that the answer is $\lfloor q/w\rfloor$ modulo 2. Each question distinguishes literal
bit values from option positions. The seven questions are evaluated in parallel and
their selected bits form one index. Values above 99 remain in the evaluation rather
than being clipped or reassigned.

\paragraph{Ordering, scheduling, and provenance.}
The reversed condition changes only dictionary insertion order. Labels stay paired
with the same descriptions. The schedule shuffles all condition/order jobs with the
fixed seed, and eight independent workers execute requests. Each job batches its
currently available questions; later rounds depend only on that job's preceding
selections. All 1,024 jobs completed without errors or retries, producing 6,863 requests
and 6,144 method outputs. Provider-reported totals are 6,982,898 input tokens,
1,773,003 output tokens, and USD~0.293282. These accounting values are not independently
audited billing and are not isolated per-method latency measurements. Request traces
exclude authorization headers. The source and case hashes were recorded before the
first request; the analysis rechecks them.

\begin{figure}[t]
\centering
\includegraphics[width=\columnwidth]{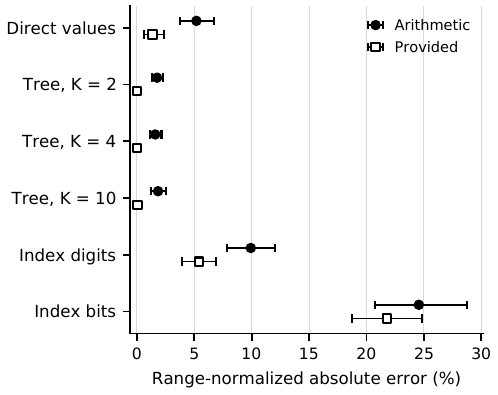}
\caption{\textbf{Error depends on representation and input condition.} Range-normalized
absolute error on the fixed affine-grid evaluation. Markers show means and horizontal
bars are 95\% family-bootstrap intervals. All transforms share 100 cells. NMAE remains
defined near zero; Table~\ref{tab:main} also reports the primary MAPE metric.}
\label{fig:affine}
\end{figure}

\section{Ordering and Affine Sensitivity}
\label{app:sensitivity}

Let $\hat y^{\uparrow},\hat y^{\downarrow}$ denote the two order-controlled outputs.
The ordering gap is $100|\hat y^{\uparrow}-\hat y^{\downarrow}|/(U-L)$, averaged
over expressions within a condition. For affine consistency, map each transformed
prediction back to grid-index space and compare it with the untransformed prediction
for the same base expression and order. Divide the difference by the 100-cell support
and multiply by 100. These quantities measure disagreement, not error against truth;
two identical but inaccurate predictions have zero disagreement.

\begin{table*}[h]
\caption{\textbf{Sensitivity as a percentage of the numerical range.} Lower values
indicate smaller order- or affine-induced output differences. Arithmetic and provided
conditions use the same expressions and transforms.}
\centering
\begin{tabular}{lrrrr}
\toprule
& \multicolumn{2}{c}{Ordering gap (\%)} & \multicolumn{2}{c}{Affine gap (\%)} \\
\cmidrule(lr){2-3}\cmidrule(lr){4-5}
Decoder & Arithmetic & Provided & Arithmetic & Provided \\
\midrule
Direct values & 1.23 & 0.78 & 6.91 & 1.82 \\
Tree ($K=2$) & 0.72 & 0.00 & 2.58 & 0.00 \\
Tree ($K=4$) & 0.88 & 0.05 & 2.09 & 0.03 \\
Tree ($K=10$) & 0.56 & 0.12 & 2.44 & 0.08 \\
Index digits & 2.02 & 2.23 & 12.70 & 7.22 \\
Index bits & 5.11 & 2.82 & 16.50 & 18.15 \\
\bottomrule
\end{tabular}
\end{table*}

\section{Exploratory Records and Distinct Protocols}
\label{app:exploratory}

The original integer expressions are $1+1$, $2+3$, $7-4$, $3\times4$, $12/3$,
and $7+8$. Fractional expressions are $1/2$, $1/4+1/8$, $3/4-1/4$,
$3/8+5/16$, $(1/2)(1/2)$, and $7/8-1/16$. Integer candidates are 0 through 15;
fraction candidates are multiples of $1/16$ in $[0,1)$. Every reference is nonzero.
There are two option orders and two repeated requests per question. Unlike the fixed
affine-grid evaluation, decimal digits here represent the actual numerical result,
using two integer digits or four fractional digits. The independent bits represent
four-bit integers or four-bit binary fractions.

\begin{table}[h]
\caption{\textbf{Exploratory arithmetic MAPE (\%).} Six integer and six fraction
questions, with four outputs each. These are small-suite descriptive results.}
\centering\small
\begin{tabular}{lrrr}
\toprule
Decoder & Integer & Fraction & Overall \\
\midrule
Direct 16-value & 0.00 & 6.73 & 3.37 \\
Four-way intervals & 0.00 & 4.83 & 2.42 \\
Explicit sets & 0.00 & 16.96 & 8.48 \\
Decimal digits & 0.00 & 31.70 & 15.85 \\
Adaptive thresholds & 0.00 & 24.62 & 12.31 \\
Parallel thresholds & 0.00 & 22.58 & 11.29 \\
Independent bits & 23.33 & 27.21 & 25.27 \\
\bottomrule
\end{tabular}
\end{table}

Explicit-set questions list the actual grid values assigned to each bit-defined set,
avoiding a request for binary encoding itself. Adaptive threshold search maintains a
selected numerical half-interval over four rounds. Parallel threshold decoding asks
15 comparisons, normalizes each yes/no pair, applies equal-weight pooled-adjacent-violators
projection, and selects the modal grid mass with lower-value tie breaking. These
threshold point estimates differ from the midpoint-weighted expectation returned by
the separate general distribution interface.

An earlier bit-prompt control contrasts underspecified bit queries with explicit bit
weights and arithmetic rules. Across the two option orders, its MAPE decreases from
40.95--46.51\% to 20.15--24.31\%. Since the new prompt adds a computational rule,
the change cannot be attributed solely to wording clarity. The original independent
bit probe, crossed prompt control, seven-method comparison, historical recall,
supplied-value control, and changed-prompt/two-branch follow-up are separate archived
protocols. We do not pool them to estimate one method-wide accuracy.

\section{Continuous-Head Transfer Protocol}
\label{app:transfer}

\paragraph{Models and prior training.}
The baselines use Qwen3.5-4B-Base and Qwen3.5-9B-Base \citep{qwen2026base},
pinned to revision prefixes
\texttt{1001bb4d826a} and \texttt{68c46c4b3498}, respectively. Each backbone supplies
its last valid hidden vector to a LayerNorm--Linear--SiLU--Linear head with hidden
width 256. The spline head predicts 49 parameters for a 16-bin monotone rational-quadratic
CDF based on the monotone spline parameterization of \citet{durkan2019spline}.
The Beta head predicts two shape parameters, constrained above 1.001.
All four checkpoints use seed 7, rank-16 LoRA \citep{hu2021lora} with scale 32, and three epochs of
negative-log-likelihood training on 12,000 structural-causal distribution queries.
Their training task describes a complete causal mechanism and intervention; it is
not arithmetic question answering. Each query has 256 sampled outcomes, with 64
sampled columns used per optimization step. Checkpoints were selected by CDF error
on the original 3,000-query validation split, before this arithmetic comparison.
The artifact retains the exact configurations and weight hashes.

\paragraph{Matched numerical inputs.}
Every checkpoint receives all 256 expressions in both the arithmetic and provided
conditions, for 512 evaluations per head. A prompt defines the known normalization
$Y=(X-L)/(U-L)$, states that $Y\in[0,1]$, and requests its distribution for the
expression in a serialized state. The provided condition adds the same numerical
result field as in the Jev evaluation. Arithmetic prompts do not contain the answer.
Appending \texttt{Distribution:} matches the trained head interface. This additional
normalization instruction is necessary for the existing head support; the prompt
is therefore semantically matched to the numerical question, not text-identical to
the Choice request. There are no choices to permute for the continuous heads.

\paragraph{Execution and readout.}
The frozen protocol uses batch size 8, 512-token left padding, BF16 backbone
activations, and FP32 head evaluation. All distribution parameters are saved.
We reconstruct the spline in FP64 on CPU and integrate $1-F(y)$ with 8,192 trapezoidal
intervals to obtain its mean; Beta means are analytic. The numerical prediction is
$\hat X=L+(U-L)\E[Y]$, without rounding to the evaluation grid. Medians are retained
as diagnostics, but not selected case by case. Evaluation performs no gradient steps
or calibration and verifies unchanged checkpoint hashes afterward. The 4B runs use
physical GPU 0 and the 9B runs GPU 1; no other device is used. Total wall time per
checkpoint, including loading and numerical readout, is approximately 78, 47, 99,
and 71 seconds for 4B spline, 4B Beta, 9B spline, and 9B Beta, respectively.

\begin{table*}[t]
\caption{\textbf{SCM-trained continuous heads transferred to the same arithmetic suite.}
All entries are percentages. The tolerance score is the fraction with APE at most
5\% (higher is better); MAPE and range-normalized MAE are lower-is-better metrics.
There are 256 expressions per condition and head, grouped into 64 bootstrap families.}
\centering\small
\begin{tabular}{lrrrrr}
\toprule
& \multicolumn{3}{c}{Arithmetic} & \multicolumn{2}{c}{Provided} \\
\cmidrule(lr){2-4}\cmidrule(lr){5-6}
Head & Within 5\% & MAPE & NMAE & Within 5\% & MAPE \\
\midrule
4B Spline & 3.91 & 117.78 & 23.99 & 3.91 & 122.66 \\
4B Beta & 2.73 & 134.67 & 24.73 & 1.17 & 119.25 \\
9B Spline & 3.52 & 138.52 & 24.59 & 5.08 & 128.19 \\
9B Beta & 6.25 & 153.88 & 26.30 & 5.47 & 148.56 \\
\bottomrule
\end{tabular}
\end{table*}

\paragraph{Interpretation.}
Figure~\ref{fig:performance} uses a post-hoc tolerance metric with the same 10,000-replicate,
64-family bootstrap as the main study. It does not redefine the predeclared primary
endpoint or select a subset of favorable examples. The supplied-answer controls compare
direct choice and the same ten-way interval decoder; both are separately marked because
their inputs contain the target. These baselines differ in backbone, task training, and representation;
their poor transfer cannot demonstrate an inherent limitation of continuous functions
or a controlled advantage of one head architecture over another.

\section{CDF Construction and Its Limits}

For threshold positions $z_1<\cdots<z_m$, normalize each returned yes/no pair to
obtain $p_i$. Equal-weight isotonic projection solves
\begin{equation}
 \min_{0\leq f_1\leq\cdots\leq f_m\leq1}\sum_{i=1}^m(f_i-p_i)^2.
\end{equation}
Adding endpoints $f_0=0$ and $f_{m+1}=1$ gives masses $w_i=f_{i+1}-f_i$.
For support constrained to $(L,U]$ and right-closed bins, these masses define a
finite histogram. A midpoint-weighted sum approximates its mean. Neither the
projection nor the midpoint rule makes the histogram calibrated to outcomes.
The raw probabilities and their monotonicity violations remain in the artifact.

\begin{figure}[h]
\centering
\includegraphics[width=\columnwidth]{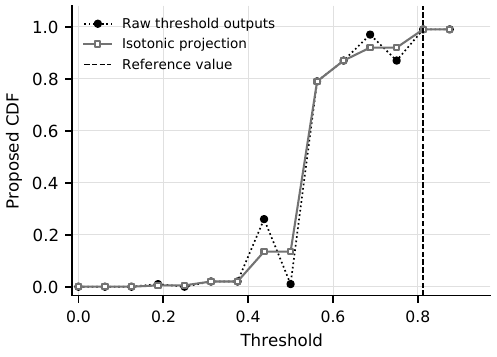}
\caption{\textbf{A shape repair does not validate uncertainty.} Raw and isotonic
threshold outputs for the archived expression $7/8-1/16$, ascending order, first
repeat. The reference is the deterministic value $13/16$. The displayed repair is
computed from the stored probabilities, not from outcome calibration data.}
\end{figure}

\section{Reproduction and Artifact Boundaries}

The manuscript build regenerates tables and vector SVG figures from the stored
metrics. SVG is the source format for every original figure; vector PDF derivatives
are included for portable TeX compilation without shell execution. The anonymous
manuscript suppresses authors, affiliation, correspondence identity, and the named
repository link. The preprint retains the author-specified affiliation and a name-only
correspondence entry.

The fixed evaluation is reproducible offline at the level of saved responses,
decoding paths, numerical metrics, and bootstrap seed. Re-querying the service may
change outputs or fail if the pinned version is unavailable. Model weights, training
data, and backend execution are not exposed. Source histories and completed negative
controls are retained, but they do not replace evaluation on additional models and
natural-language numerical tasks. The original illustration and prose were prepared
with assistance from a language-model coding agent; all quantitative results in this
manuscript are calculated from saved experimental records.

\end{document}